\documentclass[a4paper,twoside]{article}
\pdfoutput=1

\usepackage{epsfig}
\usepackage{subcaption}
\usepackage{calc}
\usepackage{amssymb}
\usepackage{amstext}
\usepackage{amsmath}
\usepackage{amsthm}
\usepackage{multicol}
\usepackage{pslatex}
\usepackage{apalike}
\usepackage[bottom]{footmisc}

\usepackage{adjustbox}
\usepackage{siunitx}
\usepackage{booktabs}
\usepackage{multirow}
\usepackage{csquotes}
\usepackage[dvipsnames]{xcolor}

\usepackage{tikzexternal}
\tikzsetexternalprefix{tikz/}

\usepackage{hyperref}
\usepackage{cleveref}
\usepackage[acronym]{glossaries}
\glsdisablehyper
\usepackage{SCITEPRESS}     
\begin{document}

\newacronym{cg}{\mbox{CG}}{Conjugate Gradient}
\newacronym{sgd}{\mbox{SGD}}{Stochastic Gradient Descent}
\newacronym{mse}{\mbox{MSE}}{Mean Square Error}
\newacronym{rmsprop}{\mbox{RMSprop}}{Root Mean Square Propagation}

\title{Training Neural Networks in Single vs.\ Double Precision}

\author{
    \authorname{
        Tomas Hrycej\sup{1},
        Bernhard Bermeitinger\sup{1}\orcidAuthor{0000-0002-2524-1850}
        and Siegfried Handschuh\sup{1}\orcidAuthor{0000-0002-6195-9034}
    }
    \affiliation{
        \sup{1}Institute of Computer Science, University of St.Gallen (HSG), St.Gallen, Switzerland
    }
    \email{\{firstname.lastname\}@unisg.ch}
}

\keywords{%
    Optimization,
    Conjugate Gradient,
    RMSprop,
    Machine Precision
}
\abstract{%
    The commitment to single-precision floating-point arithmetic is widespread in the deep learning community.
    To evaluate whether this commitment is justified, the influence of computing precision (single and double precision) on the optimization performance of the \gls{cg} method (a second-order optimization algorithm) and \gls{rmsprop} (a first-order algorithm) has been investigated.
    Tests of neural networks with one to five fully connected hidden layers and moderate or strong nonlinearity with up to 4 million network parameters have been optimized for \gls{mse}.
    The training tasks have been set up so that their \gls{mse} minimum was known to be zero.
    Computing experiments have disclosed that single-precision can keep up (with superlinear convergence) with double-precision as long as line search finds an improvement.
    First-order methods such as \gls{rmsprop} do not benefit from double precision.
    However, for moderately nonlinear tasks, \gls{cg} is clearly superior.
    For strongly nonlinear tasks, both algorithm classes find only solutions fairly poor in terms of mean square error as related to the output variance.
    \Gls{cg} with double floating-point precision is superior whenever the solutions have the potential to be useful for the application goal.
}

\onecolumn \maketitle \normalsize \setcounter{footnote}{0} \vfill

\glsresetall
\section{\uppercase{Introduction}}\label{sec:introduction}
In the deep learning community, the use of single precision computing arithmetic (the \textit{float32} format) became widespread.
This seems to result from the observation that popular first-order optimization methods for deep network training (steepest gradient descent methods) do not sufficiently benefit from a precision gain if the double-precision format is used.
This has even led to a frequent commitment to hardware without the capability of directly performing double-precision computations.
For convex minimization problems, the second-order optimization methods are superior to the first-order ones in convergence speed.
As long as convexity is given, their convergence is superlinear --– the deviation from the optimum in decimal digits decreases as fast as or faster than the number of iterations.
This is why it is important to assess whether and how far the accuracy of the second-order methods can be improved by using double precision computations (that are standard in many scientific and engineering solutions).

\section{\uppercase{Second-order optimization methods: factors depending on machine precision}}\label{sec:factors_precision}
Second-order optimization methods are a standard for numerical minimization of functions with a single local minimum.
A typical second-order method is the \gls{cg} algorithm~\cite{fletcher1964FunctionMinimizationConjugate}.
(There are also attempts to develop dedicated second-order methods, e.g., Hessian-free optimization~\cite{martens2010DeepLearningHessianfree}.)
In contrast to the first-order methods, it modifies the actual gradient in a way such that the progress made by previous descent steps is not spoiled in the actual step.
The algorithm is stopped if the gradient norm is smaller than some predefined small constant.
\Gls{cg} has the property if previous descent steps have been optimal in their descent direction, that is if a precise minimum has been reached in this direction.
This is reached by a one-dimensional optimization subroutine along the descent direction, called \emph{line search}.
Line search successively maintains three points along this line, the middle of which has a lower objective function value than the marginal ones.
The minimization is done by shrinking the interval embraced by the three points.
The stopping rule of line search consists in specifying the width of this interval at which the minimum has been reached with sufficient precision.
The precision at which this can be done is limited by the machine precision.
Second-order methods may suffer from insufficient machine precision in several ways related to the imprecision of both gradient and objective function value computation:
\begin{itemize}
    \item
        The lack of accuracy of the gradient computation may lead to distorted descent direction.
    \item
        It may also lead to a premature stop of the algorithm since the vanishing norm can be determined only with limited precision.
    \item
        It may lead to wrong embracing intervals of line search (e.g., with regard to the inequalities between the three points).
    \item
        It may also lead to a premature stop of line search if the interval width reduction no longer succeeds.
\end{itemize}

There are two basic parameters to control the computation of the \gls{cg} optimization:
There is a threshold for testing the gradient vector for being close to zero.
This parameter is usually set at a value close to the machine precision, for example, \num{e-7} for single-precision and \num{e-14} for double-precision.
The only reason to set this parameter to a higher value is to tolerate a less accurate solution for economic reasons.
Another threshold defines the width of the interval embracing the minimum of line search.
Following the convexity arguments presented in~\cite{press_NumericalRecipes2ndEdArt_1992}, this threshold should not be set to a lower value than a square root of machine precision to prevent useless line search iterations hardly improving the minimum.
Its value is above \num{e-4} for single precision and \num{3e-8} for double precision~\cite{press_NumericalRecipes2ndEdArt_1992}.
There may be good reasons to use even higher thresholds:
low values lead to more line search iterations per one gradient iteration.
Under overall constraints of computing resources such as limiting the number of function calls, it may be more efficient to accept a less accurate line search minimum, gaining additional gradient iterations.
The experiments have shown that the influence of the tolerance parameter is surprisingly low.
There is a weak preference for large tolerances.
This is why the tolerance of \num{e-1} has been used for both single and double precision in the following experiments.
The actual influence in the typical neural network optimization settings can be evaluated only experimentally.
To make the results interpretable, it is advantageous to use training sets with known minimums.
They can be generated in the following way:
\begin{enumerate}
    \item
        Define a neural network with specified (e.g., random) weights.
    \item
        Define a set of input values.
    \item
        Determine the output values resulting from the forward pass of the defined network.
    \item
        Set up a training set consisting of the defined input values and the corresponding computed results.
\end{enumerate}
This training set is guaranteed to have a minimum error of zero.

\section{\uppercase{Controlling the extent of nonlinearity}}\label{sec:extent_nonlinearity}
It can be expected that the influence of machine precision depends on the problem.
The most important aspect is the problem size.
Beyond this, the influence can be different for relatively easy problems close to convexity (nearly linear mappings) on one hand and strongly non-convex problems (nonlinear mappings).
This is why it is important to be able to generate problems with different degrees of non-convexity.
The tested problems are feedforward networks with one, two, or five consecutive hidden layers, and a linear output layer.
The bias is ignored in all layers.
All layers are fully connected.
For the hidden layers, the symmetric sigmoid with unity derivative at $ x = 0 $ has been used initially as the activation function:
\begin{equation}\label{eq:sym_sigmoid_unityderiv}
    \frac{ 2 }{ 1 + e^{ -2 x}} - 1
\end{equation}

Both input pattern values and network weights used for the generation of output patterns are drawn from the uniform distribution.
The distribution of input values is uniform on the interval $ \left< a, b \right> = \left< -1, 1 \right> $ whose mean value is zero and variance is
\begin{equation}\label{eq:uniform_var}
    \frac{ b^3 - a^3 }{ 3 \left( b - a \right)} = \frac{1}{3}
\end{equation}

To control the degree of nonlinearity during training data generation, the network weights are scaled by a factor $ c $ so that they are drawn from the uniform distribution $\left< -c, c \right>$.
The variance of the product of an input variable $ x $ and its weight $ w $ is
\begin{equation}\label{eq:wx_var}
\begin{aligned}
    \frac{1}{4c}
    \int^c_{-c}
    \int^1_{-1}
    {\left(wx \right)}^2 dx dw
    &=
    \frac{1}{4c}
    \int^c_{-c} w^2
    \frac{2}{3} dw
    \\
    &=
    \frac{1}{6c}
    \int^c_{-c}
    w^2 dw
    \\
    &=
    \frac{1}{6c}
    \frac{2c^3}{3}
    =
    \frac{c^2}{9}
\end{aligned}
\end{equation}

For large $ N $, the sum of $ N $ products $ w_i x_i $ converges to the normal distribution with variance $ N \frac{c^2}{9}$ and standard deviation $ \sqrt{N} \frac{c}{3} $.
This sum is the argument of the sigmoid activation function of the hidden layer.
The degree of nonlinearity of the task can be controlled by a normalized factor $ d $ such that $ c = \frac{d}{\sqrt{N}}$, resulting in the standard deviation $ \frac{d}{3} $ of the sigmoid argument.
In particular, it can be evaluated which share of activation arguments is larger than a certain value.
Concretely, about twice the standard deviation or more is expected to occur in \SI{5}{\percent} of the cases.

If a sigmoid function in form~\cref{eq:sym_sigmoid_unityderiv} is directly used, its derivative is close to zero with values of input argument $ x $ approaching $ 2 $.
For normalizing factor $ d = 2 $, the derivative is lower than $ 0.24 $ at \SI{5}{\percent} of the cases, compared to the derivative of unity for $ x = 0 $.
For normalizing factor $ d = 4 $, the derivative is lower than $ 0.02 $ at \SI{5}{\percent} of the cases.
The vanishing gradient problem is a well-known obstacle to the convergence of the minimization procedure.
The problem can be alleviated if the sigmoid is supplemented by a small linear term defining the guaranteed minimal derivative.
\begin{equation} \label{eq:sym_sigmoid_unityderiv_lin}
    \left( 1 - h \right)
    \left(\frac{2}{1 + e^{-2 x}} - 1 \right)
    + hx
\end{equation}

For such sigmoid without saturation with $ h = 0.05 $, the derivatives are more advantageous:
For normalizing factor $ d = 2 $, the derivative is lower than $ 0.28 $ at \SI{5}{\percent} of the cases, compared to the derivative of unity for $ x = 0 $.
For normalizing factor $ d = 4 $, the derivative is lower than $ 0.07 $ at \SI{5}{\percent} of the cases.

This activation function~\cref{eq:sym_sigmoid_unityderiv_lin} is used for the strongly nonlinear training scenarios.

\section{\uppercase{Comparison with RMSprop}}\label{sec:comparison_rmsprop}
In addition to comparing the performance of the \gls{cg} algorithm (as a representative of second-order optimization algorithms) with alternative computing precisions, it is interesting to know how competitive the \gls{cg} algorithm is compared with other popular algorithms (mostly first-order).
Computing experiments with the packages \emph{TensorFlow/Keras}~\cite{tensorflowdevelopers2022TensorFlow,chollet2015Keras} and various default optimization algorithms suggest a clear superiority of one of them: \gls{rmsprop}~\cite{hinton_NeuralNetworksMachineLearning_2012}.
In fact, this algorithm was the only one with performance comparable to \gls{cg}\@.
Other popular algorithms such as \gls{sgd} were inferior by several orders of magnitude.
This makes a comparison relatively easy: \gls{cg} is to be contrasted to \gls{rmsprop}.
\Gls{rmsprop} modifies the simple fixed-step-length gradient descent by adding a scaling factor $ \sqrt{d_{t, i} } $ depending on the iteration $ t $ and the network parameter element index $ i $.
\begin{equation}\label{eq:rmsprop}
\begin{aligned}
    w_{t + 1, i}
    & =
        w_{t,i}
        -
        \frac{c}{\sqrt{d_{t, i} }}
        \frac{\partial E \left( w_{t, i} \right)}{\partial w_{t,i}} \\
    d_{t,i}
    & =
        gd_{t-1, i} + \left( 1 - g \right)
        {\left(\partial \frac{ E \left( w_{t - 1, i} \right) }{ \partial w_{t - 1, i}} \right)}^2
\end{aligned}
\end{equation}
This factor corresponds to the weighted norm of the derivative sequence of the given parameter vector element.
In this way, it makes the steps of parameters with small derivatives larger than those with large derivatives.
If the convex error function is imagined to be a \enquote{bowl}, it makes a lengthy oval bowl more circular and thus closer to a normalized problem.
It is a step toward the normalization done by \gls{cg} but only along the axes of individual parameters, not their linear combinations.

\section{\uppercase{Computing results}}\label{sec:computing_results}
The \gls{cg} method \cite{press_NumericalRecipes2ndEdArt_1992} with \emph{Brent} line search has been implemented in \emph{C} and applied to the following computing experiments.
It has been verified by form published in~\cite{nocedal2006NumericalOptimization} (implemented in the scientific computing framework \emph{SciPy}~\cite{virtanen2020SciPyFundamentalAlgorithms}), with the line search algorithm from~\cite{wolfe1969ConvergenceConditionsAscent}.

Amendments of \gls{cg} dedicated to optimize neural networks have also been proposed: \emph{K-FAC} \cite{martens2015OptimizingNeuralNetworksa}, \emph{EKFAC} \cite{george2021FastApproximateNatural}, or \emph{K-BFGS} \cite{ren2022KroneckerfactoredQuasiNewtonMethods}.
They may possibly improve the performance of \gls{cg} in comparison to \gls{rmsprop}.

All training runs are optimized with a limit of \num{3 000} epochs for tasks with up to four million parameters.
Smaller tasks had around \num{30 000}, \num{300 000}, and 1 million parameters.
The configuration of the reported largest networks with one or five hidden layers can be seen in~\cref{tab:networks}.
The mentioned epoch limit cannot be satisfied exactly since the \gls{cg} algorithm always stops after a complete conjugate gradient iteration and thus a complete line search could consist of multiple function/gradient calls.
The number of gradient calls is generally variable per one optimization iteration, however, during the experiments they were always evaluated as often as forward passes.

\begin{table*}
    \caption{The two different network configurations with four million parameters.}\label{tab:networks}
    \centering
    \begin{tabular}{lrrrrr}
        \toprule
        Name & \# Inputs & \# Outputs & \# Hidden Layers & Hidden Layer Size & \# parameters \\
        \midrule
        4mio-1h & \multirow{2}{*}{\num{4 000}} & \multirow{2}{*}{\num{2 000}} & 1 & 680 & \num{4 080 000} \\
        4mio-5h & & & 5 & 510 & \num{4 100 400} \\
        \bottomrule
    \end{tabular}
\end{table*}

The concept of an \emph{epoch} in both types of experiments corresponds to one optimization step through the full training data with exactly one forward and one backward pass.
For \gls{cg}, the number of forward/backward passes can vary independently and the number of \emph{equivalent epochs} is adapted accordingly: a forward pass alone (as used in line search) counts as one equivalent epoch while forward and backward pass (as used in gradient computations) counts as two.
This is conservative with regard to the advantage of the \gls{cg} algorithm since the ratio between the computing effort for backward and forward passes is between one and two, depending on the number of hidden layers.
In a conservative \emph{C} implementation, equivalent epochs were roughly corresponding to the measured computing time.
For the reported result, a \emph{TensorFlow/Keras} implementation, a meaningful computing time comparison has not been possible because of the different usage of both methods: \gls{rmsprop} as an optimized built-in method against embedding \gls{cg} via \emph{SciPy} which adds otherwise unnecessary data operations.

The computing expense relationship between single and double precision depends on the hardware and software implementations.
A customary notebook, where these computations have been performed in \emph{C}, there was no difference between both machine precisions.
Other configurations may require more time for double precision, by a factor of up to four.

With this definition, \gls{cg} can be handicapped by up to \SI{33}{\percent} in the following reported results.
In the further text, \emph{epochs} refer to \emph{equivalent epochs}.

Using higher machine precision with first-order methods, including \gls{rmsprop}, brings about no significant effect.
Rough steps in the direction of the gradient, modified by equally rough scaling coefficients, whose values are strongly influenced by user-defined parameters such as $ g $ in~(\cref{eq:rmsprop}) do not benefit from high precision.
In none of our experiments with both precisions, there was a discernible advantage by double-precision.
This is why the following comparison is shown for
\begin{itemize}
    \item
        single and double precision \gls{cg} method and
    \item
        single precision \gls{rmsprop}.
\end{itemize}

To assess the optimization performance, statistics over large numbers of randomly generated tasks would have to be performed.
However, resource limitations of the two implementation frameworks do not allow such a consequent approach for large networks with at least millions of parameters.
And it is just such large networks for which the choice of the optimization method is important.
This is why several tasks of progressively growing size have been generated for networks with depths of one, two, and five hidden layers, one for each combination of size and depth.
Every task has been run with single and double precision.
In the following, only the results for the largest network size are reported since no significant differences have been observed for smaller networks.
The networks with two hidden layers behave like a compromise between a single hidden layer and five hidden layers and are thus also omitted from the presentation.

Random influences affecting such well-defined algorithms as \gls{cg} are to be taken into account when interpreting the differences between the attempts.
The convexity condition can be (and frequently is) violated so that better algorithms may be set to a suboptimal search path for some time.
For the final result to be viewed as better, the difference must be significant.
Intuitively, differences by an order of magnitude or more can be taken as significant while factors of three, two, or less are not so --- they may turn to the opposite if provided some additional iterations.

To make the minimum \gls{mse} reached practically meaningful, the results are presented as a quotient $ Q $ of the finally attained \gls{mse} and the training set output variance.
In this form, the quotient corresponds to the complement of the well-known coefficient of determination $ R^2 $, which is the ratio of the variability explained by the model to the total variability of output patterns.
The relationship between both is
\begin{equation}\label{eq:q_factor}
    Q
    =
    \frac{
        \text{MSE}}
    {
        \text{Var} \left( y \right)
    }
    =
    1 - R^2
\end{equation}
If $ Q $ is, for example, \num{0.01}, the output can be predicted by the trained neural network model with an MSE corresponding to \SI{1}{\percent} of its variance.

\subsection{Moderately nonlinear problems}\label{sec:moderately_nonlinear}
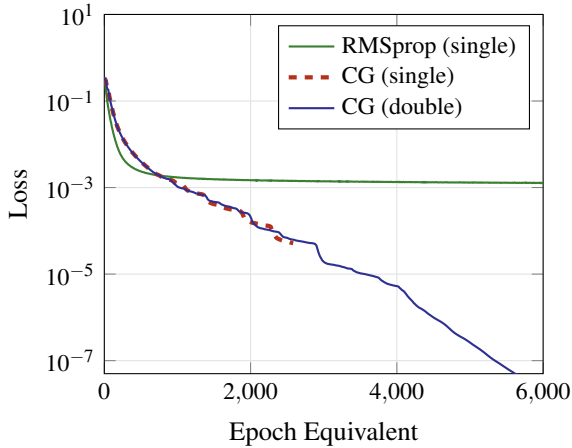
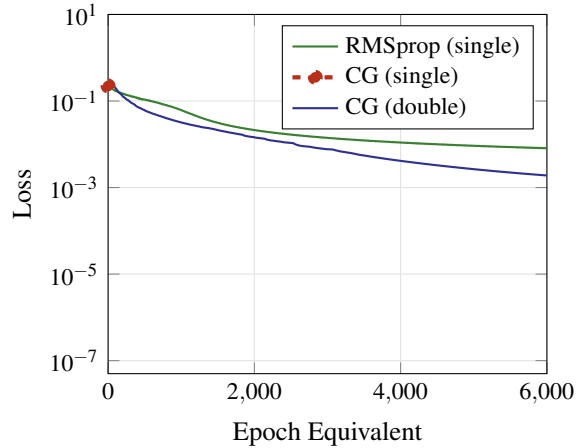
\begin{figure*}[t]
    \centering
    \begin{subfigure}[t]{.49\linewidth}
        \centering
        \tikzsetnextfilename{4mio_1_moderate.pdf}
        \begin{tikzpicture}
        \begin{semilogyaxis} [
            default,
            legend pos=north east,
            xlabel={Epoch Equivalent},
            ylabel={Loss},
            ymax=10e0,
            ymin=10e-8,
            enlargelimits=false,
            enlarge y limits={abs value=2,lower},
            width=.95\textwidth,
        ]
            \addplot [
              rmsprop64
            ] table [
                x=x, y=4mio_1_moderate_RMSPROP_float32
            ] {data/4mio_1_moderate.data};
            \addlegendentry{RMSprop (single)}

            \addplot [
              cg32,
            ] table [
                x=x, y=4mio_1_moderate_CG_float32
            ] {data/4mio_1_moderate.data};
            \addlegendentry{CG (single)}

            \addplot [
              cg64
            ] table [
                x=x, y=4mio_1_moderate_CG_float64
            ] {data/4mio_1_moderate.data};
            \addlegendentry{CG (double)}
        \end{semilogyaxis}
        \end{tikzpicture}
        \caption{One hidden layer (\emph{4mio-1h}), moderate nonlinearity.}\label{fig:h1mod}
    \end{subfigure}
    \hfill
    \begin{subfigure}[t]{.49\linewidth}
        \centering
        \tikzsetnextfilename{4mio_5_moderate.pdf}
        \begin{tikzpicture}
        \begin{semilogyaxis} [
            default,
            legend pos=north east,
            xlabel={Epoch Equivalent},
            ylabel={Loss},
            ymax=10e0,
            ymin=10e-8,
            enlargelimits=false,
            enlarge y limits={abs value=2,lower},
            width=.95\textwidth,
            ]
            \addplot [
                rmsprop64,
                ] table [
                    x=x, y=4mio_5_moderate_RMSPROP_float32
                ] {data/4mio_5_moderate.data};
            \addlegendentry{RMSprop (single)}

            \addplot [
                cg32,
                mark=*,
                ] table [
                    x=x, y=4mio_5_moderate_CG_float32
                ] {data/4mio_5_moderate.data};
            \addlegendentry{CG (single)}

            \addplot [
                cg64,
                ] table [
                    x=x, y=4mio_5_moderate_CG_float64
                ] {data/4mio_5_moderate.data};
            \addlegendentry{CG (double)}
        \end{semilogyaxis}
        \end{tikzpicture}
        \caption{Five hidden layers (\emph{4mio-5h}), moderate nonlinearity.}\label{fig:h5mod}
    \end{subfigure}
    \vspace{0.2cm}
    \caption{The largest one/five-hidden-layer networks (four million parameters) with moderate nonlinearity, loss progress (in log scale) in dependence on the number of epochs.}\label{fig:mod}
\end{figure*}

Networks with weights generated with nonlinearity parameter $ d = 2 $ (see~\cref{sec:extent_nonlinearity}) can be viewed as \emph{moderately} nonlinear.

\Cref{fig:mod} shows the attained loss measure defined in~\cref{eq:q_factor} as it develops with equivalent epochs.
The genuine minimum is always zero.
The plots correspond to \gls{rmsprop} in single precision as well as to \gls{cg} in single and double precision.
The networks considered have one (\cref{fig:h1mod}) and five (\cref{fig:h5mod}) hidden layers.

Optimization results with shallow (single-hidden-layer) neural networks have shown that the minimum \gls{mse} (known to be zero due to the task definitions) can be reached with a considerable precision of \num{e-6} to \num{e-20} even for the largest networks.
While double-precision computation is not superior to single precision for smaller networks (in a range of one order of magnitude), the improvements for the networks with one and four million parameters are in the range of two to six orders of magnitude.
The optimization progress for the largest network, comparing the dependence on the number epoch equivalents for single and double-precision arithmetic is shown in~\cref{fig:h1mod}.

With both precisions, \gls{cg} exhibits superlinear convergence property: between epochs \num{1 000} and \num{6 000}, the logarithmic plot is approximately a straight line.
So every iteration leads to an approximately fixed multiplicative gain of precision of the minimum actually reached.
The single-precision computation, however, stops after \num{2 583} epochs (\num{156} iterations) because line search can't find a better result given the low precision boundary.
In other words, the line search in single precision is less efficient.
This would be an argument in favor of double precision.

For the network with a single hidden layer, the \gls{cg} algorithm is clearly superior to \gls{rmsprop}.
The largest network attains a minimum error precision better by five orders of magnitude, and possibly more when increasing the number of epochs.
The reason is the superlinear convergence of \gls{cg} obvious from (\cref{fig:h1mod}).
It is interesting that in the initial phase, \gls{rmsprop} descends faster, quickly reaching the level that is no longer improved in the following iterations.

However, with a growing number of hidden layers, the situation changes.
For five hidden layers, single-precision computations lag behind by two orders of magnitude.
The reason for this lag is different from that observed with the single-hidden-layer tasks:
the single-precision run is prematurely stopped after less than five \gls{cg} iterations, because of no improvement in the line search.
This is why the line of the loss for single-precision \gls{cg} can hardly be discerned in~\cref{fig:h5mod}.
By contrast, the double-precision run proceeds until the epoch limit is reached (\num{655} iterations).

As seen in~\cref{fig:h5mod}, the superlinear convergence property with the double-precision computation is satisfied at least segment-wise: a faster segment until about \num{1 000} epoch equivalents and a slower segment from epoch \num{4 000}.
Within each segment, the logarithmic plot is approximately a straight line.
(Superlinearity would have to be rejected if the precision gain factors would be successively slowing down, particularly within the latter segment.)

For \gls{rmsprop}, a lag behind the \gls{cg} can be observed with five hidden layers, but the advantage of \gls{cg} is minor --- one order of magnitude.
However, the plot suggests that \gls{cg} has more potential for further improvement if provided additional resources.
Once more, \gls{rmsprop} exhibits fast convergence in the initial optimization phase followed by weak improvements.

\subsection{Strongly nonlinear problems}\label{sec:strongly_nonlinear}
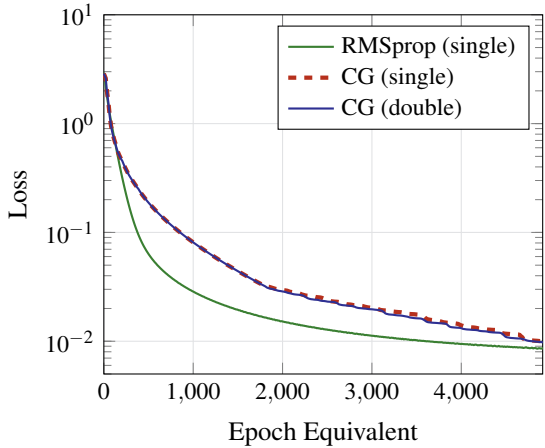
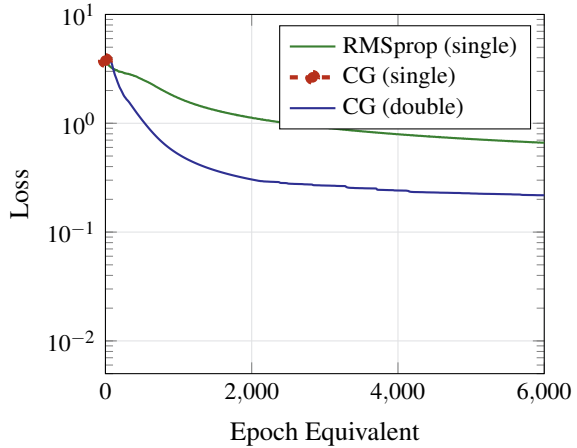
\begin{figure*}[t]
\centering
    \begin{subfigure}[t]{.49\linewidth}
        \centering
        \tikzsetnextfilename{4mio_1_strong.pdf}
        \begin{tikzpicture}
        \begin{semilogyaxis} [
            default,
            legend pos=north east,
            xlabel={Epoch Equivalent},
            ylabel={Loss},
            ymax=10e0,
            ymin=10e-3,
            enlargelimits=false,
            enlarge y limits={abs value=2,lower},
            width=.95\textwidth,
        ]
            \addplot [
              rmsprop64
            ] table [
                x=x, y=4mio_1_strong_RMSPROP_float32
            ] {data/4mio_1_strong.data};
            \addlegendentry{RMSprop (single)}

            \addplot [
              cg32,
            ] table [
                x=x, y=4mio_1_strong_CG_float32
            ] {data/4mio_1_strong.data};
            \addlegendentry{CG (single)}

            \addplot [
              cg64
            ] table [
                x=x, y=4mio_1_strong_CG_float64
            ] {data/4mio_1_strong.data};
            \addlegendentry{CG (double)}
        \end{semilogyaxis}
        \end{tikzpicture}
        \caption{One hidden layer (\emph{4mio-1h}), strong nonlinearity.}\label{fig:h1str}
    \end{subfigure}%
    \begin{subfigure}[t]{.49\linewidth}
        \centering
        \tikzsetnextfilename{4mio_5_strong.pdf}
        \begin{tikzpicture}
        \begin{semilogyaxis} [
            default,
            legend pos=north east,
            xlabel={Epoch Equivalent},
            ylabel={Loss},
            ymax=10e0,
            ymin=10e-3,
            enlargelimits=false,
            enlarge y limits={abs value=2,lower},
            width=.95\textwidth,
            ]
            \addplot [
                rmsprop64,
                ] table [
                    x=x, y=4mio_5_strong_RMSPROP_float32
                ] {data/4mio_5_strong.data};
            \addlegendentry{RMSprop (single)}

            \addplot [
                cg32,
                mark=*,
                ] table [
                    x=x, y=4mio_5_strong_CG_float32
                ] {data/4mio_5_strong.data};
            \addlegendentry{CG (single)}

            \addplot [
                cg64,
                ] table [
                    x=x, y=4mio_5_strong_CG_float64
                ] {data/4mio_5_strong.data};
            \addlegendentry{CG (double)}
        \end{semilogyaxis}
        \end{tikzpicture}
        \caption{Five hidden layers (\emph{4mio-5h}), strong nonlinearity.}\label{fig:h5str}
    \end{subfigure}
    \vspace{0.2cm}
    \caption{The largest one/five-hidden-layer networks (four million parameters) with strong nonlinearity, loss progress (in log scale) in dependence on the number of epochs.}\label{fig:str}
\end{figure*}

Mapping tasks generated with nonlinearity parameter $ d = 4 $ imply \emph{strong} nonlinearities in the sigmoid activation functions with \SI{5}{\percent} of activations having an activation function derivative of less than \num{0.02}.
With a linear term avoiding saturation (\cref{eq:sym_sigmoid_unityderiv_lin}), this derivative grows to \num{0.07}, a still very low value compared to the unity derivative in the central region of the sigmoid.
The results are shown in~\cref{fig:str} in the structure analogical to~\cref{fig:mod}.
With CG, the parameter optimization of networks of various sizes with one hidden layer and two hidden layers shows no significant difference between single and double-precision computations.
The attainable accuracy of the minimum has been, as expected, worse than for moderately nonlinear tasks but still fairly good: almost \num{e-5} for a single hidden layer and \num{e-2} for two hidden layers.

The relationship between both \gls{cg} and \gls{rmsprop} is similar for single-hidden-layer networks (\cref{fig:h1str}) --- \gls{cg} is clearly more efficient.
The attainable precision of error minimum is, as expected, worse than for moderately nonlinear tasks.

For five hidden layers, a similar phenomenon as for moderately nonlinear tasks can be observed: the single-precision computation stops prematurely because line search fails to find an improved value (see~\cref{fig:h5str}).
The minimum reached is in the same region as the initial solution.

The convergence of both algorithms (\gls{cg} and \gls{rmsprop}) is not very different with five hidden layers --- superiority of \gls{cg} is hardly significant.
Here, the superlinear convergence of \gls{cg} is questionable.
The reason for this may be a lack of convexity of the \gls{mse} with multiple hidden layers and strongly nonlinear relationships between input and output.
It is important to point out that the quality of the error minimum found is extraordinarily poor:
the \gls{mse} is about \SI{10}{\percent} of the output variance.
This is \SI{30}{\percent} in terms of standard deviation, which may be unacceptable for many applications.

\section{\uppercase{Summary and discussion}}\label{sec:summary}
With the \gls{cg} optimization algorithm, double precision computation is superior to single-precision in two cases:
\begin{enumerate}
    \item
        For tasks relatively close to convexity (single hidden layer networks with moderate nonlinearity), the optimization progress with double-precision seems to be faster due to a smaller number of epochs necessary to reach a line search minimum with a given tolerance.
        This allows the algorithm to perform more \gls{cg} iterations with the same number of epochs.
        However, since both single and double precision have the superlinear convergence property, the gap can be bridged by allowing slightly more iterations with single precision to reach a result equivalent to that of double precision.
    \item
        For difficult tasks with multiple hidden layers and strong nonlinearities, a more serious flaw of single-precision computation occurs:
        a premature stop of the algorithm because of failing to find an objective function improvement by line search.
        This may lead to unacceptable solutions.
\end{enumerate}
In summary, it is advisable to use double precision with the second-order methods.

The \gls{cg} optimization algorithm (with double precision computation) is superior to the first-order algorithm \gls{rmsprop} in the following cases:
\begin{enumerate}
    \item
        Tasks with moderate nonlinearities.
        The advantage of \gls{cg} is large for shallow networks and less pronounced for deeper ones.
        Superlinear convergence of \gls{cg} seems to be retained also for the latter group.
    \item
        Tasks with strong nonlinearities modeled by networks with a single hidden layer.
        Also here, superlinear convergence of \gls{cg} can be observed.
\end{enumerate}
For tasks with strong nonlinearities and multiple hidden layers, both \gls{cg} and \gls{rmsprop} (which has been by far the best converging method from those implemented in the popular \textit{TensorFlow/Keras} frameworks) show very poor performance.
This is documented by the square error attained, whose minimum is known to be zero in our training examples.
In practical terms, such tasks can be viewed as \enquote{unsolvable} because the forecast error is too large in relation to the output variability --- the model gained does not really explain the behavior of the output.

The advantage of \gls{rmsprop}, if any, in some of the strongly nonlinear cases is not very significant (factors around two).
By contrast, for tasks with either moderate nonlinearity or shallow networks, the \gls{cg} method is superior.
In these cases, the advantage of \gls{cg} is substantial (sometimes several orders of magnitude).
So in the typical case where the extent of nonlinearity of the task is unknown, \gls{cg} is the safe choice.

It has to be pointed out that tasks with strong nonlinearities in individual activation functions are, strictly speaking, intractable by any local optimization method.
Strong nonlinearities sum up to strongly non-monotonous mappings.
But square errors of non-monotonous mappings are certain to have multiple local minima with separate attractors.
For large networks of sizes common in today's data science, the number of such separate local minima is also large.
This reduces the chance of finding the global minimum to a practical impossibility, whichever optimization algorithms are used.
So the cases in which the \gls{cg} shows no significant advantage are just those \enquote{hopeless} tasks.

Next to the extent of nonlinearity, the depth of the network is an important category where the alternative algorithms show different performances.
The overall impression that the advantage of the \gls{cg} method over \gls{rmsprop} shrinks with the number of hidden layers, that is, with the depth of the network, may suggest the conjecture that it is not worth using \gls{cg} with necessarily double-precision arithmetic for currently preferred deep networks~\cite{heaton_IanGoodfellowYoshuaBengioAaron_2018}.

However, the argument has to be split into two different cases:
\begin{enumerate}
    \item
        networks with fully connected layers
    \item
        networks containing special layers, in particular convolutional ones.
\end{enumerate}

In the former case, the question is how far it is useful to use multiple fully connected hidden layers at all.
Although there are theoretical hints that in some special tasks, deep networks with fully connected layers may provide a more economical representation than those with shallow architectures~\cite{montufar_NumberLinearRegionsDeepNeural_2014} or~\cite{delalleau_ShallowVsDeepSumProductNetworks_2011}, the systematic investigation of~\cite{bermeitinger2019RepresentationalCapacityDeep} has disclosed no usable representational advantage of deep networks.
In addition to it, deep networks are substantially harder to train and thus exploit their representational potential.
This can also be seen in the results presented here.
Networks with five hidden layers, although known to have a zero error minimum, have not been able to be trained to a square error of less than \SI{10}{\percent} of the output variability.
Expressed in standard deviation, the standard deviation of the output error is more than \SI{30}{\percent} of the standard deviation of the output itself.
These \SI{30}{\percent} do not correspond to noise inherent to the task (whose error minimum is zero on the training set) but to the error caused by the inability of local optimization methods to find a global optimum.
This is a rather poor forecast.
In the case of the output being a vector of class indicators, the probability of frequently confusing the classes is high.
In this context, it has to be pointed out that no exact methods exist for finding a global optimum of nonconvex tasks of sizes typical for data science with many local minima.
The global optimization of such tasks is an NP-complete problem with solution time exponentially growing with the number of parameters.
This documents the infeasibility of tasks with millions of parameters.

\paragraph{Limitation to fully connected networks}
The conjectures of the present work cannot be simply extrapolated to networks containing convolutional layers --- this investigation was concerned only with fully connected networks.
The reason for this scope limitation is that it is difficult to select a meaningful prototype of a network with convolutional layers, even more one with a known error minimum --- the architectures with convolutional layers are too diversified and application-specific.
So the question is which optimization methods are appropriate for training deep networks with multiple convolutional layers but a low number of fully connected hidden layers (maybe a single one).
This question cannot be answered here, but it may be conjectured that convolutional layers are substantially easier to train than fully connected ones, for two reasons:
\begin{enumerate}
    \item
        Convolutional layers have only a low number of parameters (capturing the dependence within a small environment of a layer unit).
    \item
        The gradient with regard to convolutional parameters tends to be substantially larger than that of fully connected layers since it is a sum over all unit environments within the convolutional layer.
        In other words, convolutional parameters are \enquote{reused} for all local environments that make their gradient grow.
\end{enumerate}

This suggests a meaningful further work: to find some sufficiently general prototypes of networks with convolutional layers and to investigate the performance of alternative optimization methods on them, including the influence of machine precision for the second-order methods.

\bibliographystyle{apalike}
{\small \bibliography{bibliography}}

\providecommand{\noopsort}[1]{}
\begin{thebibliography}{}

\bibitem[Bermeitinger et~al.,
  2019]{bermeitinger2019RepresentationalCapacityDeep}
Bermeitinger, B., Hrycej, T., and Handschuh, S. (2019).
\newblock Representational {{Capacity}} of {{Deep Neural Networks}}: {{A
  Computing Study}}.
\newblock In {\em Proceedings of the 11th {{International Joint Conference}} on
  {{Knowledge Discovery}}, {{Knowledge Engineering}} and {{Knowledge
  Management}} - {{Volume}} 1: {{KDIR}}}, pages 532--538, {Vienna, Austria}.
  {SCITEPRESS - Science and Technology Publications}.

\bibitem[Chollet et~al., 2015]{chollet2015Keras}
Chollet, F. et~al. (2015).
\newblock Keras.

\bibitem[Delalleau and Bengio,
  2011]{delalleau_ShallowVsDeepSumProductNetworks_2011}
Delalleau, O. and Bengio, Y. (2011).
\newblock Shallow vs. {{Deep Sum-Product Networks}}.
\newblock In {\em Advances in {{Neural Information Processing Systems}}},
  volume~24. {Curran Associates, Inc.}

\bibitem[Fletcher and Reeves, 1964]{fletcher1964FunctionMinimizationConjugate}
Fletcher, R. and Reeves, C.~M. (1964).
\newblock Function minimization by conjugate gradients.
\newblock {\em The Computer Journal}, 7(2):149--154.

\bibitem[George et~al., 2021]{george2021FastApproximateNatural}
George, T., Laurent, C., Bouthillier, X., Ballas, N., and Vincent, P. (2021).
\newblock Fast {{Approximate Natural Gradient Descent}} in a
  {{Kronecker-factored Eigenbasis}}.

\bibitem[Heaton, 2018]{heaton_IanGoodfellowYoshuaBengioAaron_2018}
Heaton, J. (2018).
\newblock Ian {{Goodfellow}}, {{Yoshua Bengio}}, and {{Aaron Courville}}:
  {{Deep}} learning.
\newblock {\em Genet Program Evolvable Mach}, 19(1):305--307.

\bibitem[Hinton, 2012]{hinton_NeuralNetworksMachineLearning_2012}
Hinton, G. (2012).
\newblock Neural {{Networks}} for {{Machine Learning}}.

\bibitem[Martens, 2010]{martens2010DeepLearningHessianfree}
Martens, J. (2010).
\newblock Deep learning via {{Hessian-free}} optimization.
\newblock In {\em Proceedings of the 27th {{International Conference}} on
  {{International Conference}} on {{Machine Learning}}}, {{ICML}}'10, pages
  735--742, {Madison, WI, USA}. {Omnipress}.

\bibitem[Martens and Grosse, 2015]{martens2015OptimizingNeuralNetworksa}
Martens, J. and Grosse, R. (2015).
\newblock Optimizing {{Neural Networks}} with {{Kronecker-factored Approximate
  Curvature}}.

\bibitem[Mont{\'u}far et~al.,
  2014]{montufar_NumberLinearRegionsDeepNeural_2014}
Mont{\'u}far, G.~F., Pascanu, R., Cho, K., and Bengio, Y. (2014).
\newblock On the {{Number}} of {{Linear Regions}} of {{Deep Neural Networks}}.
\newblock In Ghahramani, Z., Welling, M., Cortes, C., Lawrence, N.~D., and
  Weinberger, K.~Q., editors, {\em Advances in {{Neural Information Processing
  Systems}} 27}, pages 2924--2932. {Curran Associates, Inc.}

\bibitem[Nocedal and Wright, 2006]{nocedal2006NumericalOptimization}
Nocedal, J. and Wright, S.~J. (2006).
\newblock {\em Numerical Optimization}.
\newblock Springer Series in Operations Research. {Springer}, {New York}, 2nd
  ed edition.

\bibitem[Press et~al., 1992]{press_NumericalRecipes2ndEdArt_1992}
Press, W.~H., Teukolsky, S.~A., Vetterling, W.~T., and Flannery, B.~P. (1992).
\newblock {\em Numerical Recipes in {{C}} (2nd Ed.): The Art of Scientific
  Computing}.
\newblock {Cambridge University Press}, {USA}.

\bibitem[Ren et~al., 2022]{ren2022KroneckerfactoredQuasiNewtonMethods}
Ren, Y., Bahamou, A., and Goldfarb, D. (2022).
\newblock Kronecker-factored {{Quasi-Newton Methods}} for {{Deep Learning}}.

\bibitem[{TensorFlow~Developers}, 2022]{tensorflowdevelopers2022TensorFlow}
{TensorFlow~Developers} (2022).
\newblock {{TensorFlow}}.
\newblock Zenodo.

\bibitem[Virtanen et~al., 2020]{virtanen2020SciPyFundamentalAlgorithms}
Virtanen, P., Gommers, R., Oliphant, T.~E., Haberland, M., Reddy, T.,
  Cournapeau, D., Burovski, E., Peterson, P., Weckesser, W., Bright, J.,
  {\noopsort{walt}}{van der Walt}, S.~J., Brett, M., Wilson, J., Millman,
  K.~J., Mayorov, N., Nelson, A. R.~J., Jones, E., Kern, R., Larson, E., Carey,
  C.~J., Polat, {\.I}., Feng, Y., Moore, E.~W., VanderPlas, J., Laxalde, D.,
  Perktold, J., Cimrman, R., Henriksen, I., Quintero, E.~A., Harris, C.~R.,
  Archibald, A.~M., Ribeiro, A.~H., Pedregosa, F., {\noopsort{mulbregt}}{van
  Mulbregt}, P., and {SciPy 1.0 Contributors} (2020).
\newblock {{SciPy}} 1.0: {{Fundamental}} algorithms for scientific computing in
  python.
\newblock {\em Nature Methods}, 17:261--272.

\bibitem[Wolfe, 1969]{wolfe1969ConvergenceConditionsAscent}
Wolfe, P. (1969).
\newblock Convergence {{Conditions}} for {{Ascent Methods}}.
\newblock {\em SIAM Rev.}, 11(2):226--235.

\end{thebibliography}

\end{document}